\documentclass[10pt,twocolumn,letterpaper]{article}

\usepackage{cvpr}              

\usepackage{graphicx}
\usepackage{amsmath}
\usepackage{amssymb}
\usepackage{booktabs}
\usepackage{algorithm}
\usepackage{algpseudocode}
\usepackage{multicol}
\usepackage{multirow}

\usepackage{pifont}
\newcommand{\cmark}{\ding{51}}%
\newcommand{\xmark}{\text{\ding{55}}}

\usepackage[pagebackref,breaklinks,colorlinks]{hyperref}

\usepackage[capitalize]{cleveref}
\crefname{section}{Sec.}{Secs.}
\Crefname{section}{Section}{Sections}
\Crefname{table}{Table}{Tables}
\crefname{table}{Tab.}{Tabs.}
\usepackage[normalem]{ulem}

\def\cvprPaperID{6278} 
\def\confName{CVPR}
\def\confYear{2023}

\begin{document}

\title{FlowFormer++: Masked Cost Volume Autoencoding for Pretraining Optical Flow Estimation}

\author{
Xiaoyu Shi$^{1,2*}$ \and
Zhaoyang Huang$^{1,2}$\thanks{Xiaoyu Shi and Zhaoyang Huang assert equal contributions.} \and
Dasong Li$^{1}$ \and Manyuan Zhang$^{1}$ \and
Ka Chun Cheung$^{2}$ \and
Simon See$^{2}$ \and
Hongwei Qin$^{3}$ \and
Jifeng Dai$^{4}$ \and
Hongsheng Li$^{1}$\thanks{Corresponding author: Hongsheng Li} \\ \and
$^{1}$Multimedia Laboratory, The Chinese University of Hong Kong \and
$^{2}$NVIDIA AI Technology Center \and $^{3}$SenseTime Research \and
$^{4}$Tsinghua University
}

\maketitle

\begin{abstract}
FlowFormer~\cite{huang2022flowformer} introduces a transformer architecture into optical flow estimation and achieves state-of-the-art performance. The core component of FlowFormer is the transformer-based cost-volume encoder. Inspired by the recent success of masked autoencoding (MAE) pretraining in unleashing transformers' capacity of encoding visual representation, we propose Masked Cost Volume Autoencoding (MCVA) to enhance FlowFormer by pretraining the cost-volume encoder with a novel MAE scheme.
Firstly, we introduce a block-sharing masking strategy to prevent masked information leakage, as the cost maps of neighboring source pixels are highly correlated. Secondly, we propose a novel pre-text reconstruction task, which encourages the cost-volume encoder to aggregate long-range information and ensures pretraining-finetuning consistency. We also show how to modify the FlowFormer architecture to accommodate masks during pretraining. Pretrained with MCVA, FlowFormer++ ranks 1st among published methods on both Sintel and KITTI-2015 benchmarks. Specifically, FlowFormer++ achieves 1.07 and 1.94 average end-point error (AEPE) on the clean and final pass of Sintel benchmark, leading to 7.76\% and 7.18\% error reductions from FlowFormer. FlowFormer++ obtains 4.52 F1-all on the KITTI-2015 test set, improving FlowFormer by 0.16.

\end{abstract}

\section{Introduction}
\label{sec:intro}

Optical flow is a long-standing vision task, targeting at estimating per-pixel displacement between consecutive video frames. It can provide motion and correspondence information in many downstream video problems, including video object detection~\cite{zhu2018towards,Wang_2018_ECCV,zhu2017flow}, action recognition~\cite{sun2018optical,piergiovanni2019representation,zhao2020improved}, and video restoration~\cite{kim2019deep,xu2019deep,gao2020flow,lai2017deep,chan2021basicvsr,sajjadi2018frame}.

Recently, FlowFormer~\cite{huang2022flowformer} introduces a transformer architecture for optical flow estimation and achieves state-of-the-art performance. The core of its success lies on two aspects: the ImageNet-pretrained transformer-based image encoder and the transformer-based cost-volume encoder. Notably, adopting an ImageNet-pretrained visual backbone leads to considerable performance gain over the train-from-scratch counterpart, indicating that random weight initialization hinders the learning of correspondence estimation. This naturally begs the question: can we also pretrain the transformer-based cost-volume encoder and thus further unleash its power to achieve more accurate optical flow?
\begin{figure}[!t]
    \begin{center}
    \includegraphics[width=0.9\linewidth, 
    ]{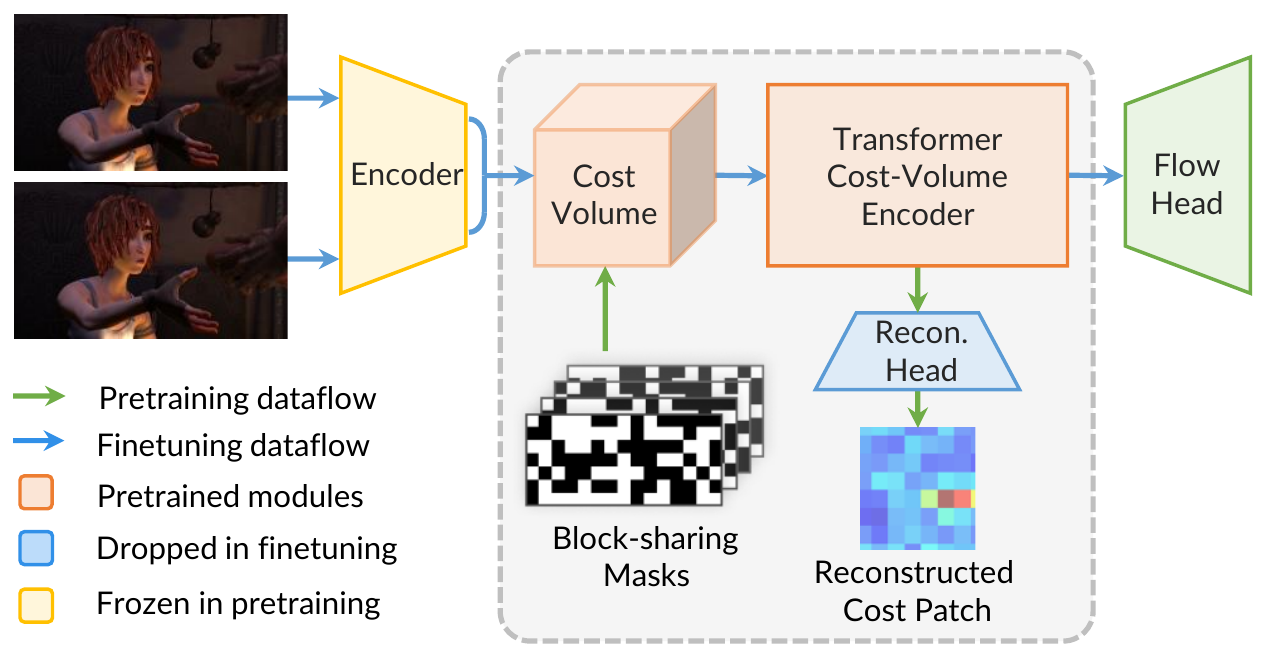}
    \end{center}
    \vspace{-0.5cm}
    \caption{$\textbf{Overview of FlowFormer++.}$ The core component of FlowFormer architecture is the transformer-based cost-volume encoder. We propose Masked Cost Volume Autoencoding to pretrain the cost-volume encoder. During pretraining, a portion of cost values are masked and the cost-volume encoder is required to reconstruct masked cost patches.} 
    \label{fig:figure1}
    \vspace{-0.4cm}
\end{figure}

In this paper, we propose masked cost-volume autoencoding (MCVA), a self-supervised pretraining scheme to enhance the cost-volume encoding on top of the FlowFormer framework. We are inspired by the recent success of masked autoencoding, such as BERT~\cite{devlin2018bert} in NLP and MAE~\cite{he2022masked} in computer vision. The key idea of masked autoencoding is masking a portion of input data, and requiring networks to learn high-level representation for masked contents reconstruction.
However, it is non-trivial to adapt the masked autoencoding strategy to learn a better cost volume encoder for optical flow estimation, because of the two following reasons. Firstly, the cost volume might contain redundancy and the cost maps (cost values between a source-image pixel to all target-image pixels) of neighboring source-image pixels are highly correlated. Randomly masking cost values, as done in other single-image pretraining methods~\cite{he2022masked}, leads to information leakage and makes the model biased towards aggregating local information. Secondly, existing masked autoencoding methods target at reconstructing masked content randomly selected from fixed locations. This suffices to pretrain general-purpose single-image encoder in other fields. However, the cost-volume encoder of FlowFormer is deeply coupled with the follow-up recurrent decoder, which demands cost information of long range at flexible locations. 

To tackle the aforementioned issues, we introduce two task-specific designs. Firstly, instead of randomly masking the cost volume, we partition source pixels into large varied-size blocks and let source pixels within the same block share a common mask pattern on their cost maps. This strategy, termed block-sharing masking, prevents the cost-volume encoder from reconstructing masked cost values by simply copying from neighboring source pixels' cost maps
Such design enfoces the cost-volume encoder to abstract useful cues from cost maps belonging to far-away source pixels, which encourages long-range information aggregation.
Secondly, to mimic the decoding process in finetuning and thus avoid pretraining-finetuning discrepancy, we propose a novel pre-text reconstruction task as shown in Fig.~\ref{fig:figure1}: small cost patches (of shape $9\times9$) are randomly cropped from the cost maps to retrieve features from the cost-volume encoder, aiming to reconstruct larger cost map patches (of shape $15\times15$) centered at the same locations. This is in line with the decoding process of FlowFormer in the finetuning stage. This pre-text task explicitly encourages the cost-volume encoder to capture 
long-range information for cost-volume encoding, which is critical for optical flow estimation. Besides, we empirically show that the image encoder, upon which the cost volume is built, should be frozen during pretraining to avoid training collapse.

In essence, the proposed masked cost-volume autoencoding (MCVA) has unique designs compared with conventional MAE methods, which encourages the cost-volume encoder 1) to construct high-level holistic representation of the cost volume, more effectively encoding long-range information, 2) to reason about occluded (\textit{i.e.}, masked) information by aggregating faithful unmasked costs, and 3) to decode task-specific feature (\textit{i.e.}, larger cost patches at required locations) to better align the pretraining process with that of the finetuning. These designs contribute to better handling of hard cases, such as noises, large-displacement motion and occlusion, for more accurate flow estimation.

To conclude, the contributions of this work are three-fold: 1) We propose the masked cost-volume autoencoding scheme to better pretrain the cost-volume encoder of FlowFormer. 2) We propose task-specific masking strategy and reconstruction pre-text task to mitigate pretraining-finetuning discrepancy, fully taking advantage of the learned representations from pretraining. 3) With the proposed pretraining technique, our proposed FlowFormer++ obtains all-sided improvements over FlowFormer, setting new state-of-the-art performance on public benchmarks.

\section{Related Work}
\label{sec:related_work}
\noindent \textbf{Optical Flow.} Compared with traditional optimization-based optical flow methods~\cite{horn1981determining,black1993framework,bruhn2005lucas,sun2014quantitative} empirically formulating flow estimation, data-driven methods~\cite{dosovitskiy2015flownet,ilg2017flownet} directly learn to estimate optical flow from labeled data.
Since FlowNet~\cite{dosovitskiy2015flownet,ilg2017flownet}, learning optical flow with neural networks presents superior performance and is still fast progressing where network architecture design becomes the key to improving optical flow accuracy. 
A series of excellent works~\cite{ranjan2017optical,sun2018pwc,sun2019models,yang2019volumetric,teed2020raft,jiang2021learning,xu2021high,jiang2021learning2,zhang2021separable,hofinger2020improving} are devoted to designing better network modules, which, indeed, introduced better inductive bias to the optical flow formulation.
For example, encoding image feature with CNNs brings locality prior, and the all-pairs 4D cost volume~\cite{teed2020raft,jiang2021learning} outperforms the coarse-to-fine cost volumes~\cite{ranjan2017optical,sun2018pwc,sun2019models} in modeling small fast-motion objects.
However, the empirical network design may always ignore some unintended cases.
Due to the success of transformers~\cite{vaswani2017attention,dai2019transformer,devlin2018bert} in image recognition~\cite{liu2021swin,dosovitskiy2020image,chu2021twins}, 
the optical flow community also tries transformers~\cite{huang2022flowformer,xu2022gmflow,sui2022craft} to further weaken the network-determined bias and learn feature relationships from data.
By replacing the handcrafted modules, \textit{i.e.}, the CNN image encoder, the cost pyramid, and the indexing-based costs retrieval, in RAFT~\cite{teed2020raft} with transformers, FlowFormer~\cite{huang2022flowformer} achieves state-of-the-art accuracy.
However, transformers are known for requiring tremendous training data to capture feature relationships~\cite{dosovitskiy2020image,he2022masked} while collecting ground-truth flows for supervised optical flow learning is expensive.
Inspired by the emerging pretraining-finetuning paradigm for vision transformers~\cite{he2022masked,gao2022convmae}, we explore to pretrain FlowFormer to capture the feature relationship for optical flow.

\noindent \textbf{Masked Autoencoding (MAE).} As a self-supervised learning technique, MAE, \eg, BERT~\cite{devlin2018bert}, achieved great success in NLP. 
Based on transformers, they mask a portion of the input tokens and require the models to predict the missing content from the reserved tokens.
Pretraining with MAE encourages transformers to build effective long-range feature relationships.
Recently, transformers also stream into the computer vision area, such as image recognition~\cite{liu2021swin,dosovitskiy2020image,chu2021twins}, video inpainting~\cite{zeng2020learning,liu2021fuseformer}, optical flow~\cite{xu2022gmflow,huang2022flowformer}, point cloud recognition~\cite{guo2021pct,zhao2021point}.
By breaking the limitations that convolution can only model local features, transformers present a significant performance gap compared to the previous counterparts.
Pretraining with MAE is also introduced to these modalities, \textit{e.g.}, image~\cite{chen2020generative,he2022masked,xie2022simmim,gao2022convmae}, video~\cite{tong2022videomae}, point cloud~\cite{yu2022point,pang2022masked}.
These works show that MAE effectively releases the transformer power and do not require extra labeled data.
FlowFormer~\cite{huang2022flowformer} presents a transformer-based cost volume encoder and achieves state-of-the-art accuracy.
In this paper, we propose the masked cost-volume autoencoding to pretrain the cost volume encoder on a video dataset, which further unleashes the power of the transformer-based cost-volume encoder.

\section{Method}
\label{sec:method}

\begin{figure*}[t!]
    \centering
    \resizebox{0.9\linewidth}{!}{
        \includegraphics[width=\linewidth, trim={5mm 42mm 5mm 25mm}, clip]{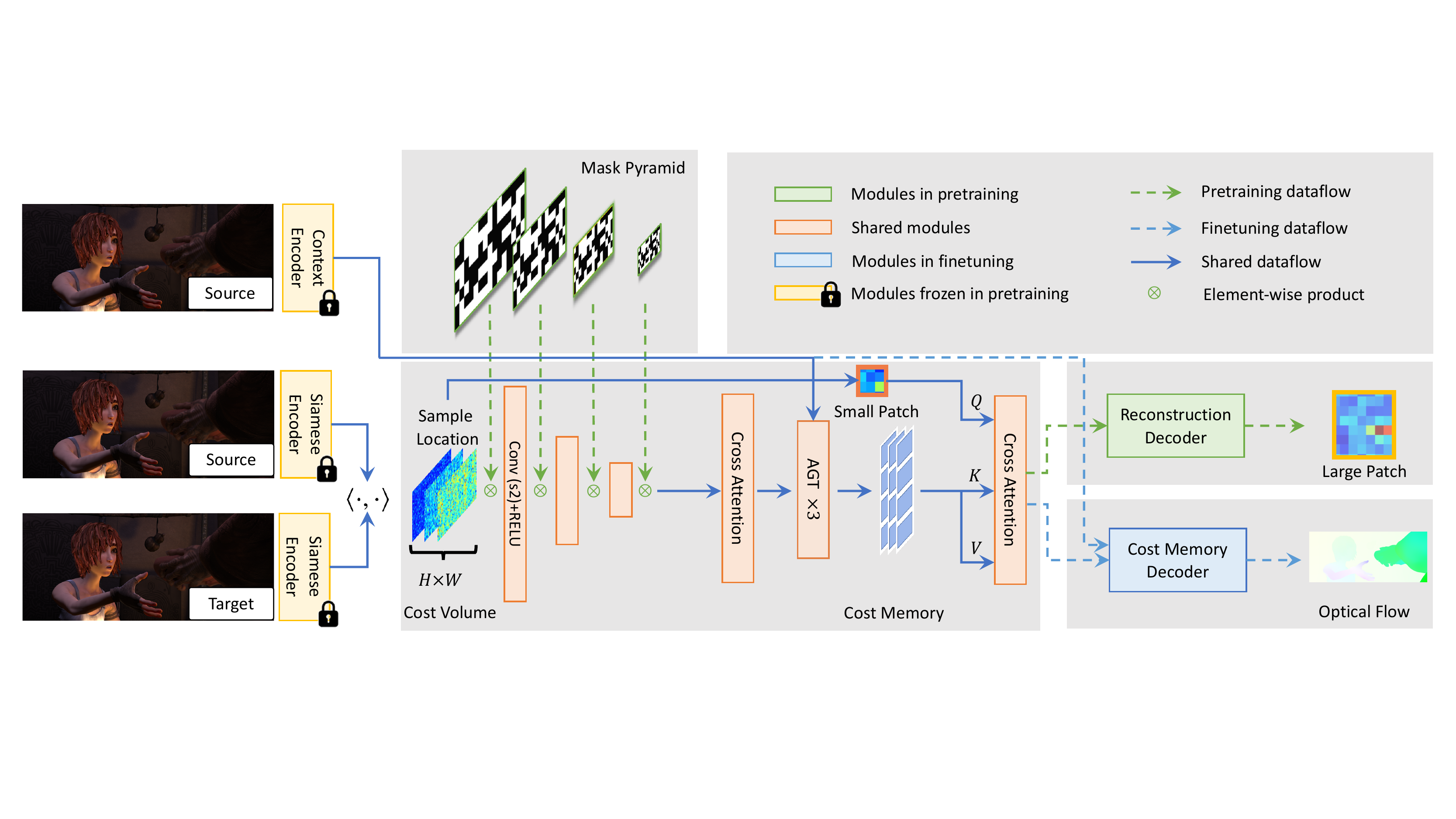}
    }
    \caption{$\textbf{Architecture of FlowFormer++.}$ During pertaining, FlowFormer++ freezes the image and context encoders, block-wisely masks the cost volume, and learns to reconstruct larger cost patches from small cost patches to pretrain cost-volume encoder. 
    In fine-tuning, FlowFormer++ uses the full cost volume, removes the reconstruction decoder, and adds the cost memory decoder to learn optical flow, which naturally falls back to the FlowFormer architecture but inherits the pretrained parameters in the cost-volume encoder.
    }
    \label{Fig: arc}
    \vspace{-0.4cm}
\end{figure*}


As presented in Fig.~\ref{Fig: arc}, we propose a masked cost-volume autoencoding (MCVA) scheme to pretrain the cost-volume encoder of FlowFormer framework for better performance.
The key of general masked autoencoding methods is to mask a portion of data and encourage the network to reconstruct the masked tokens from visible ones. 
Due to the redundant nature of the cost volume and the original FlowFormer architecture being incompatible with masks, naively adopting this paradigm to pretrain the cost-volume encoder leads to inferior performance. Our proposed MCVA tackles the challenge and conducts masked autoencoding with three key components: a proper masking strategy on the cost volume, modifying FlowFormer architecture to accommodate masks, and a novel pre-text reconstruction task supervising the pretraining process. 

In this section, we first revisit the FlowFormer architecture, and then elaborate the proposed three key designs.
We first introduce the masking strategy, dubbed as block-sharing masking, and then show the masked cost-volume tokenization that makes the cost-volume encoder compatible with masks. Coupling these two designs prevents the masked autoencoding from being hindered by information leakage in pretraining. Finally, we present the pre-text cost reconstruction task, mimicing the decoding process in finetuning to pretrain the cost-volume encoder .

\subsection{A Revisit of FlowFormer}
Given a pair of source and target images, optical flow aims at recovering pixel-level correspondences for all source pixels. FlowFormer encodes the pair of images' features with an ImageNet-pretrained Twins-SVT~\cite{chu2021twins} as $\mathbb R^{H_I\times W_I\times 3}\rightarrow \mathbb R^{H\times W\times D}$, and creates a 4D cost volume of size $H\times W\times H\times W$ by computing all-pairs feature correlations. $H_I, W_I$ and $H, W$ respectively indicate the height and width of the images and the visual feature maps. The cost volume can also be viewed as a series of cost maps of size $\mathbb R^{H\times W}$, each of which measures the similarity between one source pixel and all target pixels.

The 4D cost volume contains abundant but redundant information for optical flow estimation.
FlowFormer projects it into a latent space of size $\mathbb R^{H\times W\times K\times D}$ with a cost tokenizer.
In the latent space, each source pixel’s cost map is transformed into cost memory consisting of $K$ tokens of dimension $D$, which is a more compact representation and is further processed by a transformer-based cost encoder, dubbed as alternate-group transformer~(AGT). Finally, FlowFormer recurrently decodes the flow estimation from the cost memory with cross-attention.

FlowFormer is the first transformer architecture specifically designed for optical flow estimation, which enjoys the benefits of long-range information encoding via self-attention, but also encounters the similar problem to general vision transformers: it needs large-scale training data to model unbiased representations. 
The FlowFormer with the ImageNet-pretrained Twins-SVT backbone leads to boosted accuracy, while the same model with a train-from-scratch Twins-SVT or a shallow CNN achieve similar degraded performances, demonstrating the necessity of pretraining transformers for optical flow estimation.
However, the ImageNet can only be used for pretraining the single-image encoder and the cost-volume encoder in FlowFormer is still trained from scratch and might not converge to the optimal point.
To enable the pretraining of the cost-volume encoder to further enhance optical flow estimation, we propose the masked cost-volume autoencoding scheme.




\subsection{Block-sharing Cost Volume Masking}


A properly designed masking scheme is required to conduct autoencoding of the masked cost volume. 
For each source pixel $\mathbf x$, we need to create a binary mask ${\bf M_x} \in \{0,1\}^{H \times W}$ to its cost map ${\bf C_x} \in \mathbb{R}^{H \times W}$, where 0 indicates masking (\textit{i.e.}, removing) cost values from the masked locations.
Naturally, neighboring source pixels' cost maps are highly correlated. Randomly masking neighboring pixels' cost maps might cause information leakage, \textit{i.e.}, masked cost values might be easily reconstructed by copying the cost values from neighboring source pixels' cost maps.

To prevent such an over-simplified learning process, we propose a block-sharing masking strategy. We partition source pixels into non-overlapping blocks in each iteration. All source pixels belonging to the same block share a common mask for masked region reconstruction. In this way, neighboring source pixels are unlikely to copy each other's cost maps to over-simplify the autoencoding process. Besides, the size of block is designed to be large (height and width of blocks are of $32\sim120$ pixels) and randomly changes in each iteration, and thus encouraging the cost-volume encoder to aggregate information from long-range context and to filter noises of cost values. The details of the mask generation algorithm are provided in supplementary.

Specifically, for each source pixel's cost map, we first generate the mask map 
${\mathbf M}_{\mathbf x}^3\in \{0,1\}^{\frac{H}{8} \times \frac{W}{8}}$
at $\frac{1}{8}$ resolution, and then up-sample it $2\times$ for three times to obtain a pyramid of mask maps ${\mathbf M}_{\mathbf x}^i\in \{0,1\}^{\frac{H}{2^i} \times \frac{W}{2^i}}$, where $i \in \{0,1,2\}$, which are used for the down-sampling encoding process and will be discussed later in Sec.~\ref{sec:tokennization}.

Another key design is that, in pretraining, we {\bf freeze} the ImageNet-pretrained Twins-SVT backbone to build the cost volume from the pair of input images. Freezing the image encoder ensures the reconstruction targets (\textit{i.e.}, raw cost values) to maintain static and avoids training collapse.

\subsection{Masked Cost-volume Tokenization}
\label{sec:tokennization}

Given the above generated mask for each source pixel ${\bf x}$'s cost map,
FlowFormer adopts a two-step cost-volume tokenization before the cost encoder.
To prevent the masked costs from leaking into subsequent cost aggregation layers, 
the intermediate embeddings of the cost map need to be properly masked in the cost-volume tokenization process.
We propose the masked cost-volume tokenization, which prevents mixing up masked and visible features.
Firstly, FlowFormer patchifies the raw cost map $\mathbf C_{\mathbf x}\in \mathbb R^{H\times W}$ of each source pixel ${\mathbf x}$ (which is obtained by computing dot-product similarities between the source pixel ${\mathbf x}$ and all target pixels) via 3 stacked stride-2 convolutions. We denote the feature maps after each of the 3 convolutions as $\mathbf F_{\mathbf x}^i$, which have spatial sizes of $\frac{H}{2^i}\times \frac{W}{2^i}$ for $i \in \{0,1,2\}$.
We propose to replace the vanilla convolutions used in the FlowFormer with masked convolutions~\cite{gao2022convmae,graham2017submanifold,ren2018sbnet}:
\begin{equation}
    {\mathbf F}_{\mathbf x}^{i+1} = {\rm Conv}_{\rm stride2}\left(\mathrm{ReLU}({\mathbf F}_{\mathbf x}^i\odot \mathbf{ M}_\mathbf{x}^i)\right),
\end{equation}
where $\odot$ indicates element-wise multiplication, $i \in \{0,1,2\}$, 
and ${\mathbf F}_{\mathbf x}^0$ is the raw cost map ${\bf C_x}$. 
The masked convolutions with the three binary mask maps remove all cost features in the masked regions in pretraining.
Secondly, FlowFormer further projects the patchified cost-map features into the latent space via cross-attention. 
We thus remove the tokens in $\mathbf{F}_{\bf x}^3$ indicated by the mask map ${\mathbf M}_{\mathbf x}^3$ and then only project the remaining tokens into the latent space via the same cross-attention.
During finetuning, the mask maps are removed to utilize all cost features, which converts the masked convolution to the vanilla convolution but the pretrained parameters in the convolution kernels and cross-attention layer are maintained.

The masked cost-volume tokenization completes two tasks.
Firstly, it ensures the subsequent cost-volume encoder only processes visible features in pertaining.
Secondly, the network structure is consistent with the standard tokenization of FlowFormer and can directly be used for finetuning so that the pretrained parameters have the same semantic meanings.
After the masked cost-volume tokenization, the cost aggregation layers (\textit{i.e.}, AGT layers) take visible features as input which also don't need to be modified in finetuning. 
The latent features interact with those of other source pixels in AGT layers and are transformed to the cost memory $\mathbf T_{\mathbf x}$. We explain how to decode the cost memory to estimate flows in following section.

\subsection{Reconstruction Target for Cost Memory Decoding}
With the masked cost-volume tokenization, the cost encoder encodes the unmasked cost volume into the cost memory.
The next step is decoding and reconstructing the masked regions from the cost memory.

In this section, we formulate the pre-text reconstruction targets, which supervises the decoding process as well as aformentioned embedding and aggregation layers. We start by revisiting the dynamic positional decoding scheme of FlowFormer, and present our reconstruction targets which are highly consistent with the finetuning tasks. FlowFormer adopts recurrent flow prediction. In each iteration of the recurrent process, the flow of source pixel ${\mathbf x}$ is decoded from cost memory ${\bf T_x}$, conditioned on current predicted flow, to update the flow prediction. Specifically, current predicted corresponding location in the target image $\bf{p_x}$ is computed as $\bf{p_x=x+f(x)}$, where $\bf{f(x)}$ is current predicted flow. A local cost patch $\bf{q_x}$ is then cropped from the $9 \times 9$ window centered at ${\bf p_x}$ on the raw cost map $\bf{C_x}$. FlowFormer utilizes this local cost patch (with positional encoding) as the query feature to retrieve aggregated cost feature $\bf{c_x}$ via cross-attention operation:
\begin{equation}
\begin{aligned}
    & {\bf Q_x} = \mathrm{FFN}\left(\mathrm{FFN}(\mathbf{q_x})  + \mathrm{PE}(\mathbf p) \right), \\
    & {\bf K_x} = \mathrm{FFN}\left({\bf T_x}\right),~~{\bf V_x} = \mathrm{FFN}\left({\bf T_x}\right), \\
    & {\bf c_x} = \mathrm{Attention}({\bf Q_x}, {\bf K_x}, {\bf V_x}).
\label{eq:decoder attention}
\end{aligned}
\end{equation}

\begin{figure}[!t]
    \begin{center}
    \includegraphics[width=0.9\linewidth]{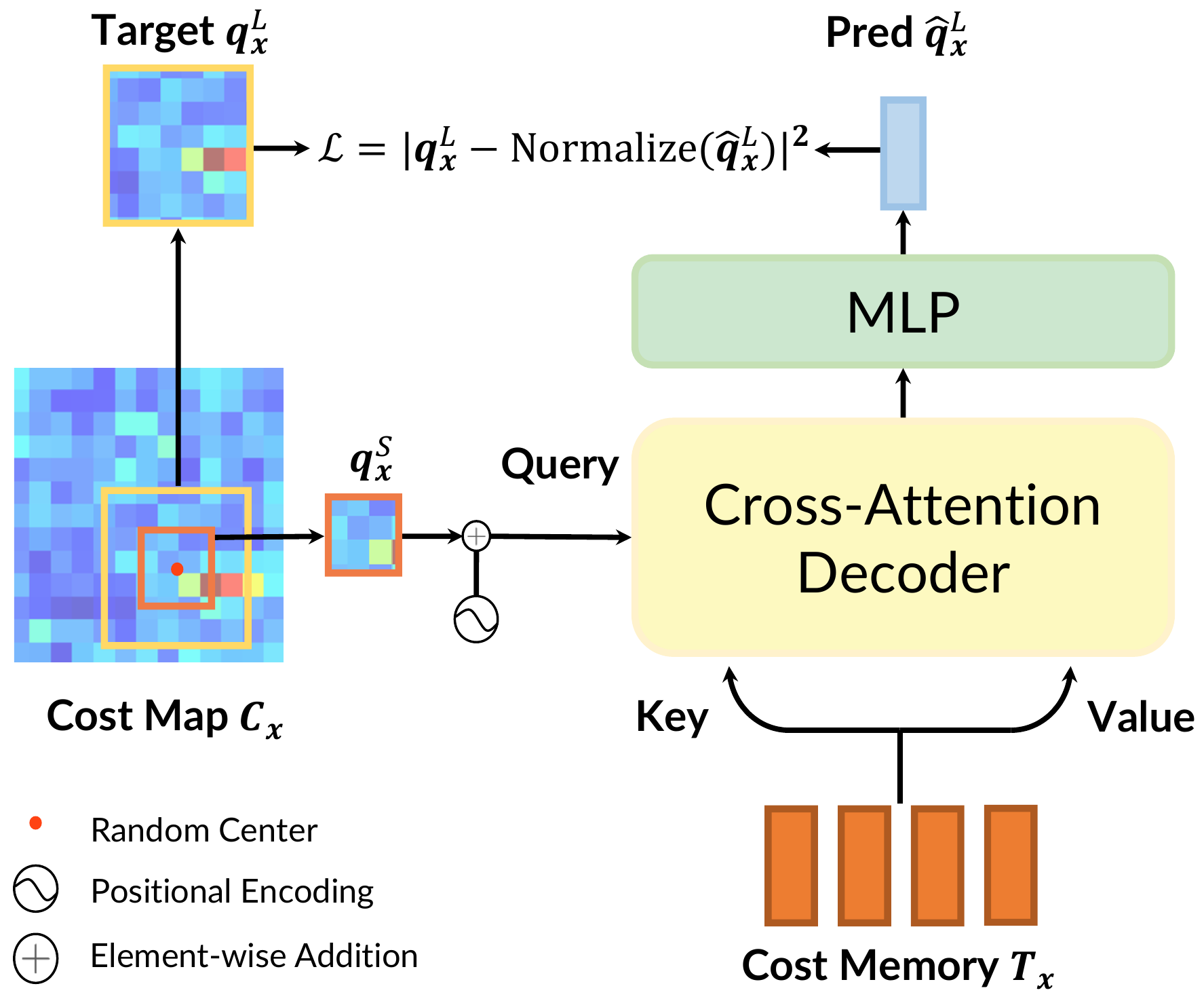}
    \end{center}
    \vspace{-0.5cm}
    \caption{$\textbf{Pre-text reconstruction for cost memory decoding.}$ For each source pixel $\bf x$ and its corresponding cost map $\bf C_x$, a small cost patch ${\bf q}_{\bf x}^S$ is randomly cropped from the cost map to retrieve features from the cost memory $\bf T_x$, aiming to reconstruct larger cost patch ${\bf q}_{\bf x}^L$ centered at the same location.}
    \label{fig:figure1}
    \vspace{-0.4cm}
\end{figure}

\noindent\textbf{Pre-text Reconstruction.}
Intuitively, ${\bf c_x}$ should contain long-range cost information for better optical flow estimation and it is conditioned on local cost patch $\bf{q_x}$, which indicates the interested location on the cost map. We design a pre-text reconstruction task in line with these two characteristics to pretrain the cost-volume encoder as shown in Fig.~\ref{fig:figure1}: small cost-map patches are randomly cropped from the cost maps to retrieve cost features from the cost memory, targeting at reconstructing larger cost-map patches centered at the same locations.

Specifically, for each source pixel ${\bf x}$, we randomly sample a location ${\bf o_x}$, which is analogous to ${\bf p_x}$ in finetuning. Taking this location as center, we crop a small cost-map patch ${\bf q}_{\bf x}^S=\mathrm{Crop}_{9\times 9}(\mathbf{C_x}, \mathbf{o_x})$ of shape $9 \times 9$. Then we perform the decoding process shown in Equation~\ref{eq:decoder attention} to obtain the cost feature ${\bf c_x}$, except that ${\bf p_X}$ and ${\bf q_x}$ are replaced by ${\bf o_x}$ and ${\bf q}_{\bf x}^S$ in pretraining, respectively. To encourage the extracted cost feature ${\bf c_x}$ to carry long-range cost information conditioned on ${\bf o_x}$, we take larger cost-map patch as supervision. Specifically, we crop another larger cost-map patch ${\bf q}_{\bf x}^L=\mathrm{Crop}_{15\times 15}(\mathbf{C_x}, \mathbf{o_x})$ of shape $15 \times 15$ centered at the same location ${\bf o_x}$. We choose a light-weight MLP as prediction head. The MLP takes as input ${\bf c_x}$ and its output ${ \mathbf{\hat q}}_{\bf x}^{L}=\mathrm{MLP}({\bf c_x})$ is supervised by normalized ${\bf q}_{\bf x}^L$. We take mean squared error (MSE) as loss function.
\begin{equation}
    \mathcal{L} = \frac{1}{|\Omega|} \sum_{{\bf x}\in \Omega} \bigg| {\bf q}_{\bf x}^{L} - \mathrm{Normalize}( {\bf{\hat{q}}}_{\bf x}^L)\bigg|^2,
\end{equation} where $\Omega$ is the set of source pixels.

\noindent \textbf{Discussion.}
The key of pretraining is to maintain consistent with finetuning, in terms of both network arthictecture and prediction target.
To this end, we keep the cross-attention decoding layer unchanged and construct inputs with the same semantic meaning (\textit{e.g.}, replacing dynamically predicted ${\bf p_x}$ with randomly sampled ${\bf o_x}$); we supervise the extracted feature ${\bf c_x}$ with long-range cost values to encourage the cost-volume encoder to aggregation global information for better optical flow estimation.
What's more, our scheme only takes an extra light-weight MLP as prediction head, which is unused in finetuning. Compared with previous methods that use a stack of self-attention layers, it is much more computationally efficient.

\begin{table*}[t]
\centering
\scriptsize
\resizebox{0.9\linewidth}{!}{
\begin{tabular}{clccccccc}
\hline
\multicolumn{1}{c}{\multirow{2}{*}{Training Data}} & \multicolumn{1}{c}{\multirow{2}{*}{Method}} & \multicolumn{2}{c}{Sintel (train)}                    & \multicolumn{2}{c}{KITTI-15 (train)}                    & \multicolumn{2}{c}{Sintel (test)}                     & \multicolumn{1}{c}{KITTI-15 (test)} \\
\cmidrule(r{1.0ex}){3-4}\cmidrule(r{1.0ex}){5-6}\cmidrule(r{1.0ex}){7-8} \cmidrule(r{1.0ex}){9-9} 
\multicolumn{1}{c}{}                               & \multicolumn{1}{c}{}                        & \multicolumn{1}{c}{Clean} & \multicolumn{1}{c}{Final} & \multicolumn{1}{c}{F1-epe} & \multicolumn{1}{c}{F1-all} & \multicolumn{1}{c}{Clean} & \multicolumn{1}{c}{Final} & \multicolumn{1}{c}{F1-all} \\ 
\hline
\multirow{3}{*}{A}                         &    Perceiver IO~\cite{jaegle2021perceiver}   & 1.81 & 2.42 & 4.98  & - & - & - & - \\ 
&    PWC-Net~\cite{sun2018pwc}   & 2.17 & 2.91 & 5.76 & - & - & - & -  \\ 
& RAFT~\cite{teed2020raft}  & 1.95 & 2.57 &  4.23 & - & - & - & - \\
\hline
\multicolumn{1}{c}{\multirow{13}{*}{C+T}}           & HD3~\cite{yin2019hierarchical} &  3.84      & 8.77      & 13.17  & 24.0  & - & - & - \\
 & LiteFlowNet~\cite{hui2018liteflownet} & 2.48 & 4.04 & 10.39 & 28.5 & - & - & -  \\
 & PWC-Net~\cite{sun2018pwc} & 2.55 & 3.93 & 10.35 & 33.7 & - & - & - \\
 & LiteFlowNet2~\cite{hui2020lightweight} & 2.24 & 3.78 & 8.97 & 25.9 & - & - & - \\
 & S-Flow~\cite{zhang2021separable}  & 1.30 & 2.59 & 4.60 & 15.9 & & & \\
 & RAFT~\cite{teed2020raft}   & 1.43 & 2.71 & 5.04 & 17.4 & - & - & - \\
 & FM-RAFT~\cite{jiang2021learning2}   & 1.29 & 2.95 & 6.80 & 19.3 & - & - & -  \\
 & GMA~\cite{jiang2021learning}   & 1.30 & 2.74 & 4.69 & 17.1 & - & - & - \\
 & GMFlow~\cite{xu2022gmflow}   & 1.08 & 2.48 & - & - & - & - & - \\
 & GMFlowNet~\cite{zhao2022global}   & 1.14 & 2.71 & 4.24 & 15.4 & - & - & - \\
 & CRAFT~\cite{sui2022craft}   & 1.27 & 2.79 & 4.88 & 17.5 & - & - & - \\
 & SKFlow~\cite{sun2022skflow}   & 1.22 & 2.46 & 4.47 & 15.5 & - & - & - \\
 
 & FlowFormer~\cite{huang2022flowformer}   & $\uline{0.94}$ & $\uline{2.33}$ & $\uline{4.09}^{\dag}$ & $\uline{14.72}^{\dag}$ & - & - & - \\
 & Ours & $\mathbf{0.90}$ & $\mathbf{2.30}$ & $\mathbf{3.93}^{\dag}$ & $\mathbf{14.13}^{\dag}$ & - & - & - \\
\hline
\multirow{15}{*}{C+T+S+K+H}                         &    LiteFlowNet2~\cite{hui2020lightweight}    &    (1.30)                       &       (1.62)                    &   (1.47)                          &      (4.8)                      &       3.48                     &         4.69                   &          7.74   \\
 &  PWC-Net+~\cite{sun2019models} & (1.71) & (2.34) &  (1.50) &  (5.3) & 3.45 &  4.60 & 7.72\\
 & VCN~\cite{yang2019volumetric} & (1.66) & (2.24) & (1.16) & (4.1) & 2.81 & 4.40 & 6.30 \\
 &  MaskFlowNet~\cite{zhao2020maskflownet} & - & - & - & - & 2.52 & 4.17 & 6.10 \\
  &  S-Flow~\cite{zhang2021separable} & (0.69) & (1.10) & (0.69) & (1.60) & 1.50 & 2.67 & $\uline{4.64}$ \\
 &  RAFT~\cite{teed2020raft} & (0.76) & (1.22) & (0.63) & (1.5) & 1.94 & 3.18 & 5.10
\\
&  RAFT*~\cite{teed2020raft} & (0.77) & (1.27) & - & - & 1.61 & 2.86 & 5.10
\\
 &  FM-RAFT~\cite{jiang2021learning2} & (0.79) & (1.70) & (0.75) & (2.1) & 1.72 & 3.60 &  6.17 \\
 &  GMA~\cite{jiang2021learning} & - & - & - & - & 1.40 & 2.88 & 5.15 \\
  &  GMA*~\cite{jiang2021learning} & (0.62) & (1.06) & (0.57) & (1.2) & 1.39 & 2.47 & 5.15 \\
  &  GMFlow~\cite{xu2022gmflow} & - & - & - & - & 1.74 & 2.90 & 9.32 \\
  &  GMFlowNet~\cite{zhao2022global} & (0.59) & (0.91) & (0.64) & (1.51) & 1.39 & 2.65 & 4.79 \\
  &  CRAFT~\cite{sui2022craft} & (0.60) & (1.06) & (0.57) & (1.20) & 1.45 & 2.42 & 4.79 \\
  &  SKFlow*~\cite{sun2022skflow} & (0.52) & (0.78) & (0.51) & (0.94) & 1.28 & 2.23 & 4.84 \\
&  FlowFormer~\cite{huang2022flowformer} & (0.48) & (0.74) & (0.53) & (1.11) & $\uline{1.16}$ & $\uline{2.09}$ & $4.68^\dag$  \\
&  Ours & (0.40) & (0.60) & (0.57) & (1.16) & $\mathbf{1.07}$ & $\mathbf{1.94}$ & $\mathbf{4.52}{^\dag}$  \\
\hline
\end{tabular}
}
\caption{\label{Tab: comparison} $\textbf{Experiments on Sintel~\cite{butler2012naturalistic} and KITTI~\cite{geiger2013vision} datasets.}$ `A' denotes the autoflow dataset. `C + T' denotes training only on the FlyingChairs and FlyingThings datasets. `+ S + K + H' denotes finetuning on the combination of Sintel, KITTI, and HD1K training sets.  * denotes that the methods use the warm-start strategy~\cite{teed2020raft}, which relies on previous image frames in a video, while other methods use two frames only. $^\dag$ denotes the result is obtained via the tile technique proposed in FlowFormer~\cite{huang2022flowformer}.
Our FlowFormer++ achieves the best generalization performance~(C+T) and ranks 1st on both the Sintel and the KITTI-15 benchmarks~(C+T+S+K+H).
}
\end{table*}

\section{Experiments}
We evaluate our FlowFormer++ on the Sintel~\cite{butler2012naturalistic} and KITTI-2015~\cite{geiger2013vision} benchmarks. We pretrain FlowFormer++ using the proposed Masked Cost-volume Autoencoding on YouTube-VOS~\cite{xu2018youtube} dataset. For the supervised finetuning, following previous works, we train FlowFormer++ on FlyingChairs~\cite{dosovitskiy2015flownet} and FlyingThings~\cite{mayer2016large}, and then respectively finetune it on the Sintel and KITTI-2015 benchmarks. FlowFormer++ obtains all-sided improvements over FlowFormer, ranking 1st on both benchmarks.

\begin{figure*}
\centering
    \resizebox{0.9\linewidth}{!}{
\setlength{\tabcolsep}{2pt}
\begin{tabular}{@{} c c c @{}}
        \includegraphics[width=.32\linewidth]{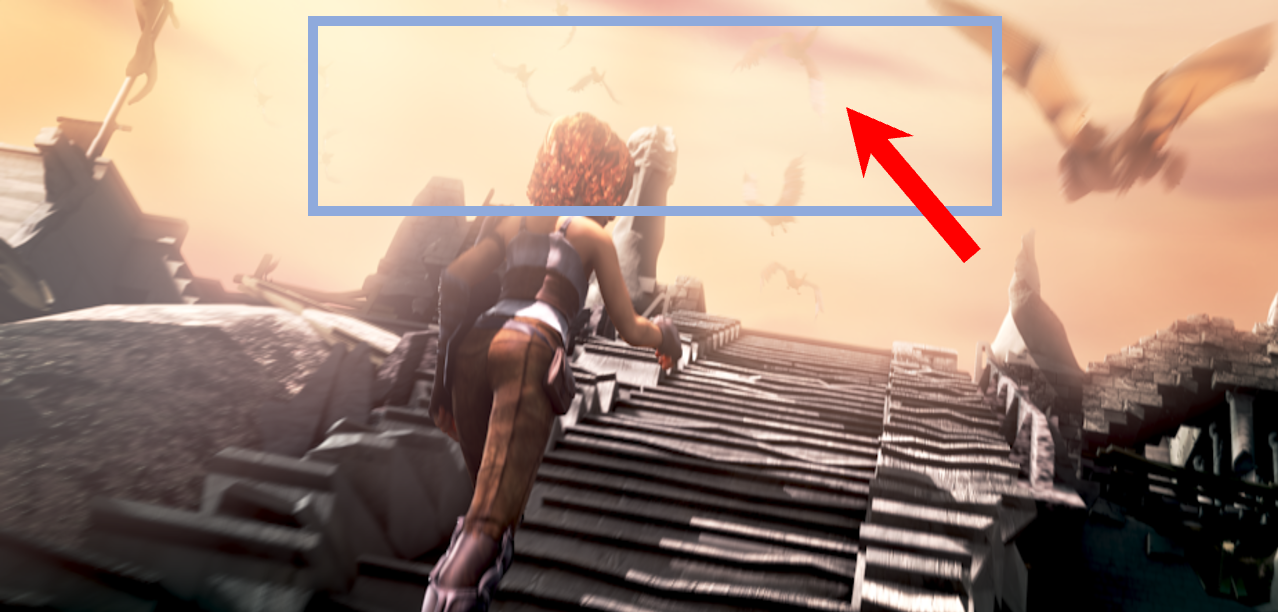} &
        \includegraphics[width=.32\linewidth]{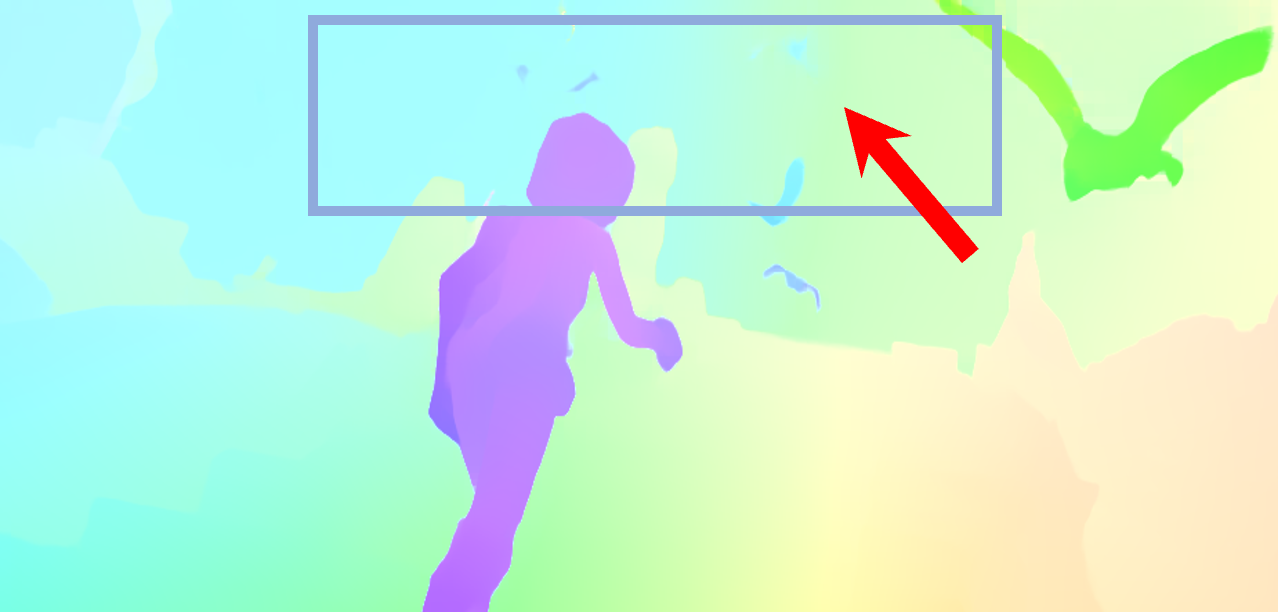} &
        \includegraphics[width=.32\linewidth]{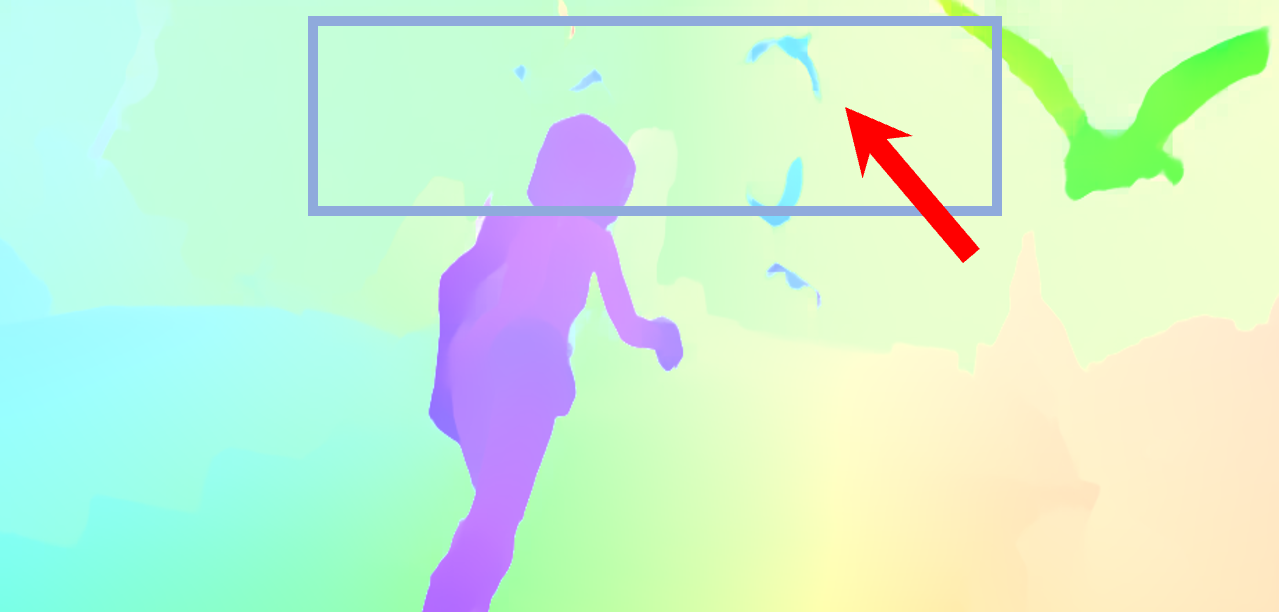}  \\
        \includegraphics[width=.32\linewidth]{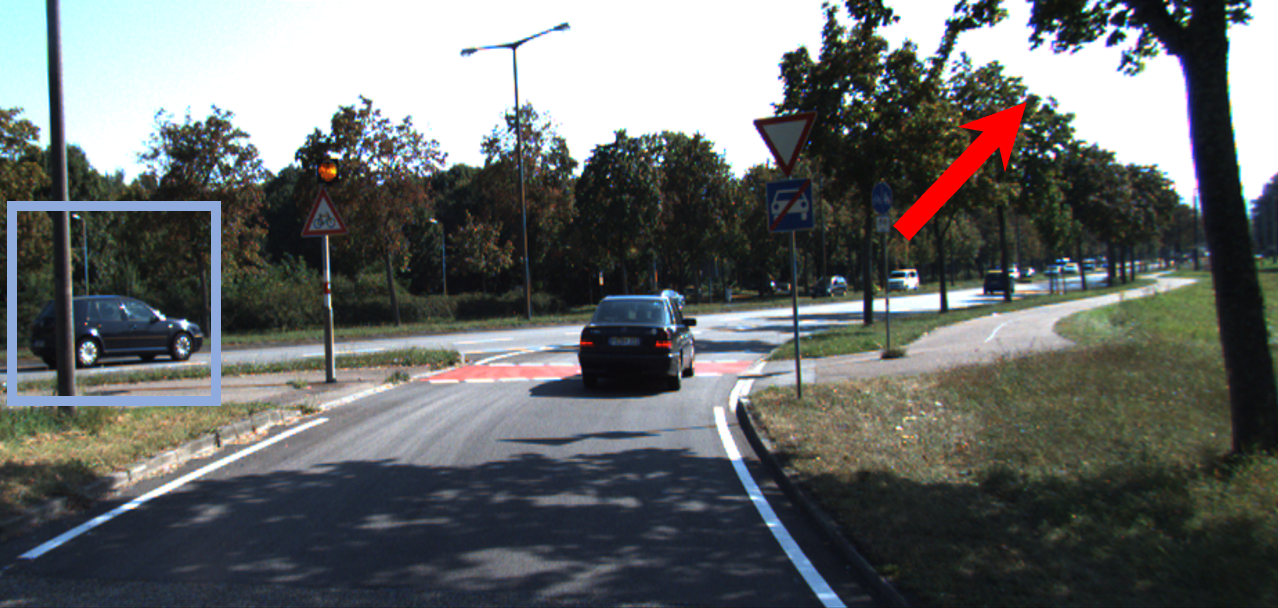} &
        \includegraphics[width=.32\linewidth]{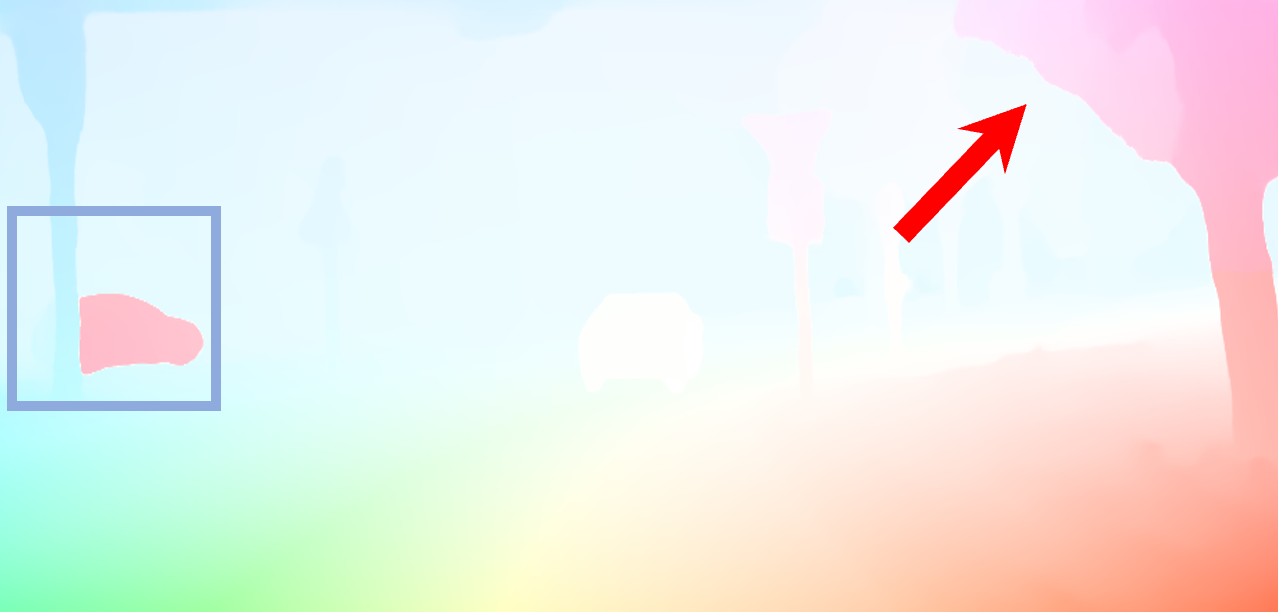} &
        \includegraphics[width=.32\linewidth]{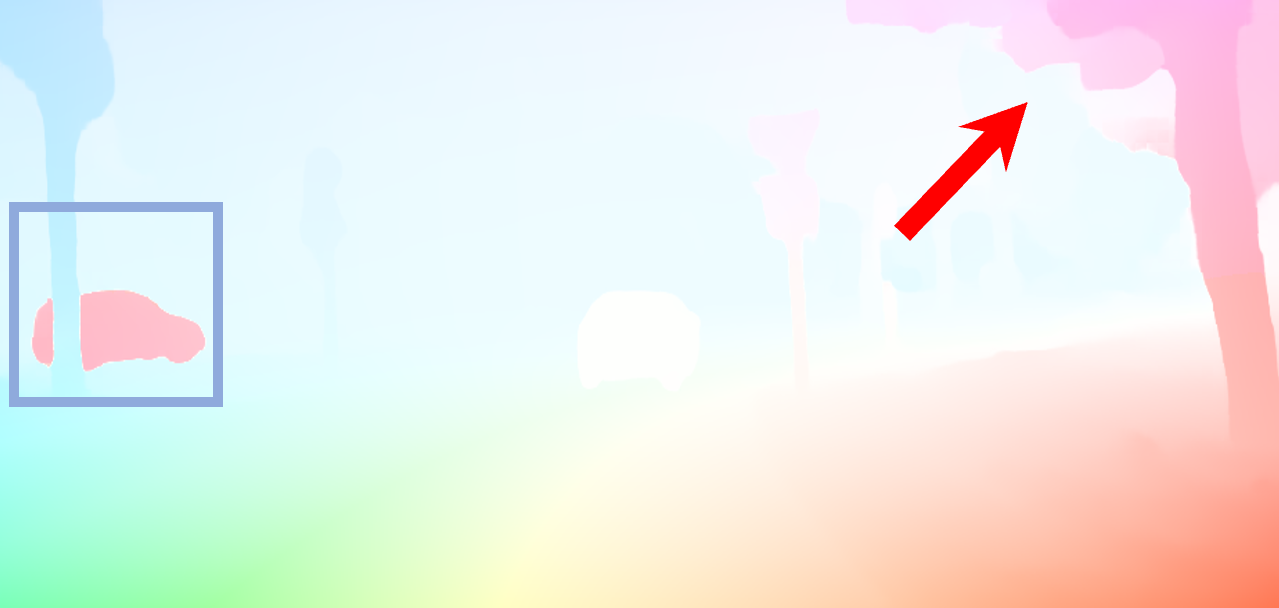} \\
        Input & FlowFormer & FlowFormer++ (Ours)  \\
    \end{tabular}
    }
    \vspace{-0.15cm}
    \caption{$\textbf{Qualitative comparison on Sintel and KITTI test sets.}$ FlowFormer++ preserves clearer details (pointed by red arrows) and maintains better global consistency (indicated by blue boxes).
    }
    \label{fig:quality}
    \vspace{-0.3cm}
\end{figure*}

\noindent \textbf{Experimental Setup.}
We adopt the commonly-used average end-point-error (AEPE) as the evaluation metric. It measures the average $l_2$ distance between predictions and ground truth. For the KITTI-2015 dataset, we additionally use the F1-all (\%) metric, which refers to the percentage of pixels whose flow error is larger than 3 pixels or over 5\% of the length of ground truth flows. YouTube-VOS is a large-scale dataset containing video clips from YouTube website. The Sintel dataset is rendered from the same movie in two passes: the clean pass is rendered with easier smooth shading and specular reflections, while the final pass includes motion blur, camera depth-of-field blur and atmospheric effects. The motions in the Sintel dataset are relatively large and complicated. The KITTI-2015 dataset constitutes of real-world driving scenarios with sparse ground truth.

\noindent \textbf{Implementation Details}.
We use the same architecture of FlowFormer for fair comparison. The image feature encoder and context feature encoder are chosen as the first two stages of ImageNet-pretrained Twins-SVT, which are frozen in pretraining for better performance. We pretrain our model on YouTube-VOS for 50k iterations with a batch size of 24. The highest learning rate is set as $5\times 10^{-4}$. During finetuning, we follow the same training procedure of FlowFormer. We train our model on FlyingChairs for 120k iterations with a batch size of 8 and on FlyingThings with a batch size of 6 (denoted as `C+T'). Then, we train FlowFormer++ by combining data from Sintel, KITTI-2015 and HD1K~\cite{kondermann2016hci} (denoted as `C+T+S+K+H') for another 120k iterations with a batch size of 6. This model is submitted to Sintel online test benchmark for evaluation. To obtain the best performance on the KITTI-2015 benchmark, we further train FlowFormer++ on the KITTI-2015 dataset for 50k iterations with a batch size of 6. The highest learning rate is set as $2.5\times 10^{-4}$ for FlyingChairs and $1.25\times 10^{-4}$ on other training sets. In both pretraining and finetuning, we choose AdamW optimizer and one-cycle learning rate scheduler. We crop images and tile predictions from all patches to obtain full-resolution flow predictions following Perceiver IO~\cite{jaegle2021perceiver} and FlowFormer.

\subsection{Quantitative Experiments}

As shown in Table~\ref{Tab: comparison}, we evaluate FlowFormer++ on the Sintel and KITII-2015 benchmarks. Following previous methods, we evaluate the generalization performance of models on the training sets of Sintel and KITTI-2015 (denoted as `C+T'). We also compare the dataset-specific accuracy of optical flow models after dataset-specific finetuning (denoted as `C+T+S+K+H'). Autoflow~\cite{sun2021autoflow} is a synthetic dataset of complicated visual disturbance, while its training code is unreleased. 

\noindent \textbf{Generalization Performance.}
The `C+T' setting in Table~\ref{Tab: comparison} reflects the generalization capacity of models. FlowFormer++ ranks 1st on both benchmarks among published methods. It achieves 0.90 and 2.30 on the clean and final pass of Sintel. Compared with FlowFormer, it achieves 4.26\% error reduction on Sintel clean pass. On the KITTI-2015 training set, FlowFormer++ achieves 3.93 F1-epe and 14.13 F1-all, improving FlowFormer by 0.16 and 0.59, respectively. These results show that our proposed MCVA promotes the generalization capacity of FlowFormer.

\noindent \textbf{Dataset-specific Performance.}
After training the FlowFormer++ in the `C+T+S+K+H' setting, we evaluate its performance on the Sintel online benchmark. It achieves 1.07 and 2.09 on the clean and final passes, a 7.76\% and 7.18\% error reduction from previous best model FlowFormer.

\begin{figure*}[t!]
    \centering
    \resizebox{0.9\linewidth}{!}{
        \includegraphics[width=0.98\linewidth, trim={0mm 0mm 0mm 0mm}, clip]{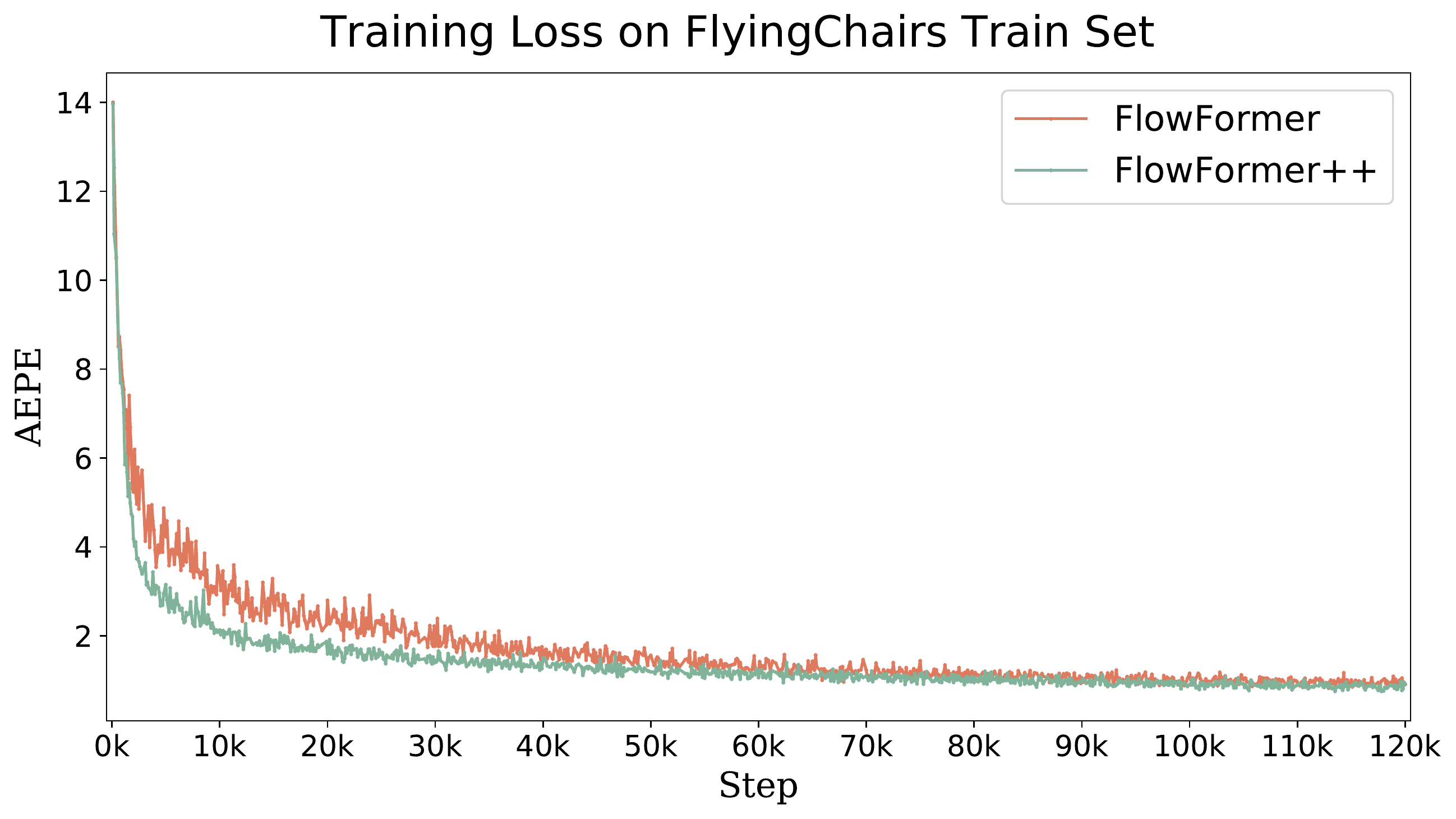}
        \includegraphics[width=\linewidth, trim={0mm 0mm 0mm 0mm}, clip]{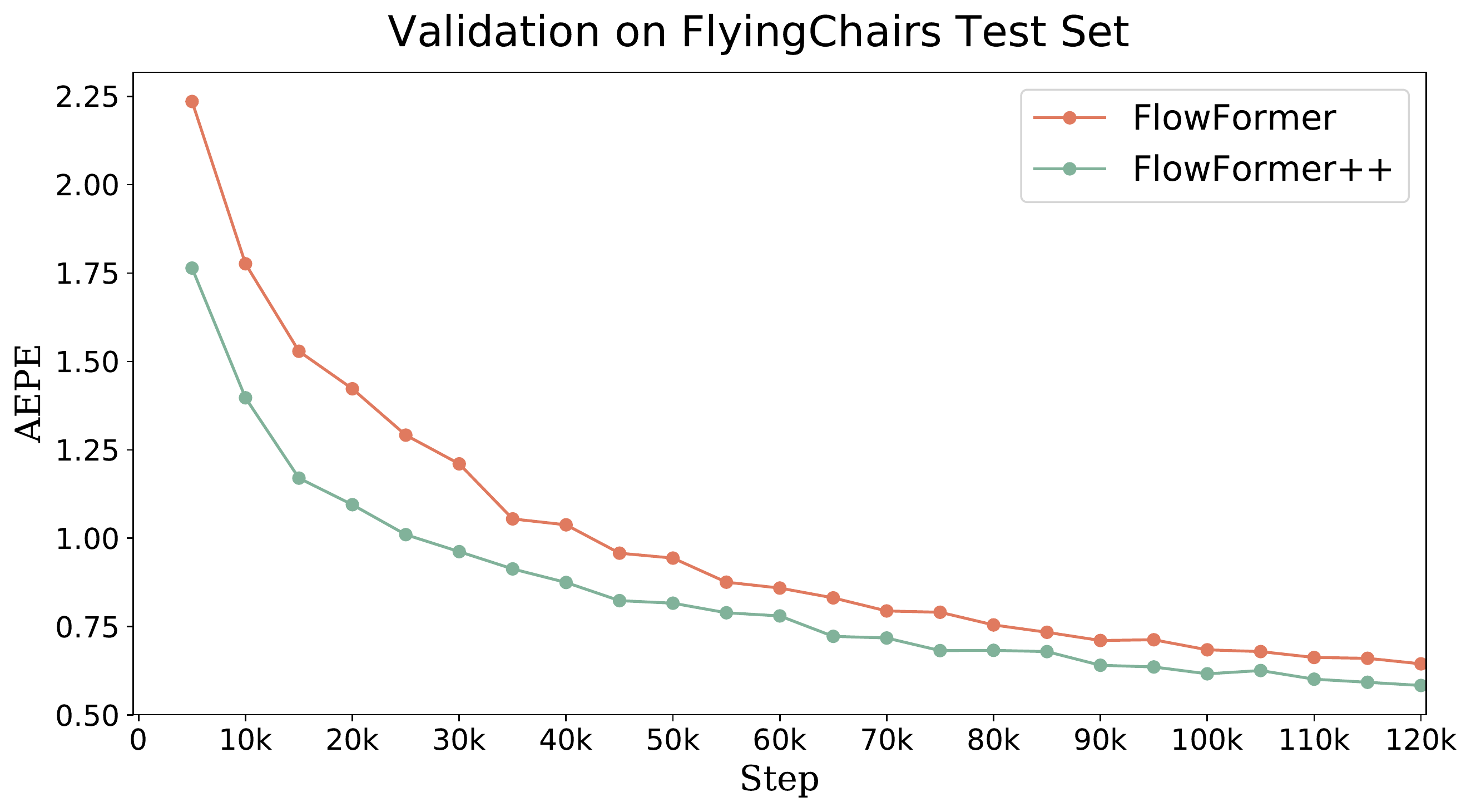}
    }
    \caption{$\textbf{Comparison on FlyingChairs.}$ FlowFormer++ converges faster and achieves lower validation error.
    }
    \label{Fig: training curve}
\end{figure*}

We further finetune FlowFormer++ on the KITTI-2015 training set after the Sintel stage and evaluate its performance on the KITTI online benchmark. FlowFormer++ achieves 4.52 F1-all, improving FlowFormer by 0.16 while also outperforming the previous best model S-Flow by 0.12.

To conclude, FlowFormer++ shows greater optical flow estimation capacity for both naturalistic non-rigid motions (Sinel) and real-world rigid scenarios (KITTI-2015). This validates that our proposed MCVA improves the FlowFormer architecture by enhancing the cost-volume encoder.

\subsection{Qualitative Experiments}
We visualize flow predictions by our FlowFormer++ and FlowFormer on Sintel and KITTI test sets in Fig.~\ref{fig:quality} to qualitatively show how FlowFormer++ outperforms FlowFormer. The red arrows highlight that FlowFormer++ preserves clearer details than FlowFormer: in the first row, FlowFormer misses the flying bird while FlowFormer++ produces clear results; in the second row, FlowFormer++ keeps the boundaries of leaves while FlowFormer generates blurry prediction. FlowFormer++ also shows greater global aggregation capacity indicated by blue boxes. In the first row, FlowFormer produces obviously inconsistent prediction on the large-area sky, while FlowFormer++ yields consistent prediction. In the second row, the black car is partially occluded by the foreground tree, which challenges the optical flow model to aggregate information in long range. FlowFormer++ generates consistent prediction for the two separated parts of the car, while FlowFormer mixes the left part of the car with background and thus produces inconsistent optical flow prediction.
\subsection{Ablation Study}
We conduct a set of ablation studies to show the impact of designs in the Masked Cost-volume Autoencoding (MCVA). All models in the experiemnts are first pretrained and then finetuned on `C+T'. We report the test results on Sintel and KITTI training sets.

\noindent \textbf{Masking Strategy.}
Masking strategy is one important design of our MCVA. As shown in Table~\ref{Tab: masking strategy}, pretraining FlowFormer with random masking already improves the performance on three of the four metrics. But the proposed block-sharing masking strategy brings even larger gain, which demonstrates the effectiveness of this design. Besides, we observe higher pretraining loss with block-sharing masking than that with random masking, validating that the block-sharing masking makes the pretraining task harder.

\noindent \textbf{Masking Ratio.}
Masking ratio influences the difficulty of the pre-text reconstruction task. We empirically find that the mask ratio of 50\% yields the best overall performance.

\noindent \textbf{Pre-text Reconstruction Design.} The conventional MAE methods aim to reconstruct input data at fixed locations and use the positional encodings as query features to absorb information for reconstruction (the first row of Table~\ref{Tab: Pre-text reconstruction design.}). To keep consistent with the dynamic positional query of FlowFormer architecture, we propose to reconstruct contents at random locations (the second row of Table~\ref{Tab: Pre-text reconstruction design.}) and additionally use local patches as query features (the third row of Table~\ref{Tab: Pre-text reconstruction design.}). The results validate the necessity of ensuring semantic consistency between pretraining and finetuning. 

\noindent \textbf{Freezing Image and Context Encoders.}
The FlowFormer architecture has an image encoder to encode visual appearance features for constructing the cost volume, and a context encoder to encode context features for flow prediction. As shown in Table~\ref{Tab: freezing encoders.}, freezing the image encoder is necessary, otherwise the model diverges. We hypothesize that the frozen image encoder ensures the reconstruction targets (\textit{i.e.}, raw cost values) to keep static. Freezing the context encoder leads to better overall performance.
\begin{table}[t]
\centering
\scriptsize
\begin{tabular}{lcccc}
\hline
 \multicolumn{1}{c}{\multirow{2}{*}{Case}} & \multicolumn{2}{c}{Sintel (train)}                    & \multicolumn{2}{c}{KITTI-15 (train)} \\
\cmidrule(r{1.0ex}){2-3} \cmidrule(r{1.0ex}){4-5}
        & \multicolumn{1}{c}{Clean} & \multicolumn{1}{c}{Final} & \multicolumn{1}{c}{F1-epe} & \multicolumn{1}{c}{F1-all} \\ 
\hline
FlowFormer & 0.94 & $\uline{2.33}$ & 4.09  & 14.72 \\
+ Random Masking & $\uline{0.93}$ & 2.35 & $\uline{4.03}$ & $\uline{14.30}$ \\ 
+ Block-sharing Masking  & $\mathbf {0.90}$ & $\mathbf {2.30}$ & $\mathbf {3.93}$ & $\mathbf{14.13}$  \\ 
\hline
\end{tabular}
\caption{\label{Tab: masking strategy} $\textbf{Masking strategy.}$ Pretraining with block-sharing masking brings greater performance gain than that with random masking.
}
\end{table}

\begin{table}[t]
\centering
\scriptsize
\begin{tabular}{ccccc}
\hline
 \multicolumn{1}{c}{\multirow{2}{*}{Masking Ratio}} & \multicolumn{2}{c}{Sintel (train)}                    & \multicolumn{2}{c}{KITTI-15 (train)} \\
\cmidrule(r{1.0ex}){2-3} \cmidrule(r{1.0ex}){4-5}
        & \multicolumn{1}{c}{Clean} & \multicolumn{1}{c}{Final} & \multicolumn{1}{c}{F1-epe} & \multicolumn{1}{c}{F1-all} \\ 
\hline
20\% & 0.97 & 2.41 & $\mathbf{3.91}$  & 14.27 \\
50\% & $\mathbf {0.90}$ & $\mathbf{2.30}$ & $\uline{3.93}$ & $\uline{14.13}$ \\ 
80\% & $\uline{0.94}$ & $\uline{2.35}$ & 4.00 & $\mathbf{14.07}$ \\ 
\hline
\end{tabular}
\caption{\label{Tab: masking ratio} $\textbf{Masking ratio.}$ The masking ratio influences the difficulty of the reconstruction task. Masking 50\% cost values yields best overall results. 
}
\vspace{-10px}
\end{table}

\begin{table}[t]
\centering
\scriptsize
\begin{tabular}{lccccc}
\hline
 {\multirow{2}{*}{Location}} &{\multirow{2}{*}{Query Feature}} 
 & \multicolumn{2}{c}{Sintel (train)}                    
 & \multicolumn{2}{c}{KITTI-15 (train)} \\
\cmidrule(r{1.0ex}){3-4} \cmidrule(r{1.0ex}){5-6}
    & \multicolumn{1}{c}{}    & \multicolumn{1}{c}{Clean} & \multicolumn{1}{c}{Final} & \multicolumn{1}{c}{F1-epe} & \multicolumn{1}{c}{F1-all} \\ 
\hline
Fixed & PE & 0.99 & $\uline{2.40}$ & 4.35 & 15.33 \\
Random & PE & $\uline{0.95}$ & 2.42 & $\uline{3.99}$ & $\uline{14.47}$ \\ 
Random & PE + Cropped patch & $\mathbf{0.90}$ & $\mathbf{2.30}$ & $\mathbf{3.93}$ & $\mathbf{14.13}$ \\ 
\hline
\end{tabular}
\caption{\label{Tab: Pre-text reconstruction design.} $\textbf{Pre-text reconstruction design.}$ Our pre-text reconstruction task leads to better performance over conventional MAE task (first row) and its improved version with random reconstruction locations (second row).
}
\end{table}

\begin{table}[t]
\centering
\scriptsize
\begin{tabular}{cccccc}
\hline
  \multicolumn{2}{c}{Freeze} 
 & \multicolumn{2}{c}{Sintel (train)}                    
 & \multicolumn{2}{c}{KITTI-15 (train)} \\
\cmidrule(r{1.0ex}){1-2} \cmidrule(r{1.0ex}){3-4} \cmidrule(r{1.0ex}){5-6}
  Image Encoder  & Context Encoder & \multicolumn{1}{c}{Clean} & \multicolumn{1}{c}{Final} & \multicolumn{1}{c}{F1-epe} & \multicolumn{1}{c}{F1-all} \\ 
\hline
\xmark & \xmark & - & - & - & - \\
\xmark & \cmark & - & - & - & - \\ 
\cmark & \xmark & $\uline{0.92}$ & $\uline{2.34}$ & $\mathbf{3.90}$ & $\uline{14.23}$ \\ 
\cmark & \cmark & $\mathbf{0.90}$ & $\mathbf{2.30}$ & $\uline{3.93}$ & $\mathbf{14.13}$ \\ 
\hline
\end{tabular}
\caption{\label{Tab: freezing encoders.} $\textbf{Freezing image and context encoders in pretraining.}$ Freezing the image encoder ensures the reconstruction targets (\textit{i.e.}, raw cost values) to maintain static, otherwise the model diverges. Freezing the context encoder leads to better overall performance.
}
\end{table}

\begin{table}[t]
\centering
\scriptsize
\begin{tabular}{lcccc}
\hline
 \multicolumn{1}{c}{\multirow{2}{*}{Methods}} & \multicolumn{2}{c}{Sintel (train)}                    & \multicolumn{2}{c}{KITTI-15 (train)} \\
\cmidrule(r{1.0ex}){2-3} \cmidrule(r{1.0ex}){4-5}
        & \multicolumn{1}{c}{Clean} & \multicolumn{1}{c}{Final} & \multicolumn{1}{c}{F1-epe} & \multicolumn{1}{c}{F1-all} \\ 
\hline
Unsupervised Baseline & $\uline{0.99}$ & $\uline{2.54}$ & $\uline{4.38}$ & $\uline{15.22}$ \\ 
MCVA (ours)  & $\mathbf {0.90}$ & $\mathbf {2.30}$ & $\mathbf {3.93}$ & $\mathbf{14.13}$  \\ 
\hline
\end{tabular}
\caption{\label{Tab: unsupervised} $\textbf{Comparisons with unsupervised methods.}$ We use the conventional unsupervised algorithm~\cite{stone2021smurf,9477059,Luo_2021_CVPR} (using photometic loss and smooth loss) to pretrain FlowFormer for comparison. Our MCVA outperforms the unsupervised counterpart.
}
\end{table}
\noindent \textbf{Comparisons with Unsupervised Methods.}
We also use conventional unsupervised methods to pretrain FlowFormer with photometric loss and smooth loss following~\cite{stone2021smurf,9477059,Luo_2021_CVPR} and then finetune it in the `C+T' setting as FlowFormer++. As shown in Table~\ref{Tab: unsupervised}, our MCVA outperforms the unsupervised counterpart for pretraining FlowFormer.

\noindent \textbf{FlowFormer++ v.s. FlowFormer on FlyingChairs.
}
We show the training and validating loss of the training process on FlyingChairs~\cite{dosovitskiy2015flownet} in Fig.~\ref{Fig: training curve}.
FlowFormer++ presents faster convergence during training and better validation loss at the end, which reveals that FlowFormer++ learns effective feature relationships during pretraining and benefits the supervised finetuning.

\section{Conclusion}
In this paper, we propose Masked Cost Volume Autoencoding (MCVA) to enhance the cost-volume encoder of FlowFormer by pretraining. We show that the naive adaptation of MAE scheme to cost volume does not work due to the redundant nature of cost volumes and the incurred pretraining-finetuning discrepancy. We tackle these issues with a specially designed block-sharing masking strategy and the novel pre-text reconstruction task. These designs ensure semantic integrity between pretraining and finetuning and encourage the cost-volume to aggregate information in a long range. Experiments demonstrate clear generalization and dataset-specific performance improvements.
{\small
\bibliographystyle{ieee_fullname}
\bibliography{egbib}
}

\end{document}